\documentclass[final]{cvpr}

\usepackage{times}
\usepackage{epsfig}
\usepackage{graphicx}
\usepackage{subfigure}
\usepackage{amsmath}
\usepackage{amssymb}
\usepackage{multirow}
\usepackage{makecell}
\usepackage{amsfonts}
\usepackage{amssymb}

\usepackage[pagebackref=true,breaklinks=true,colorlinks,bookmarks=false]{hyperref}

\begin{document}

\title{AsymmNet: Towards ultralight convolution neural networks \\using asymmetrical bottlenecks}

\author{
Haojin Yang\thanks{Equal contribution. 
This work is done when Haojin Yang and Yuncheng Zhao are with Alibaba Cloud.}\\
Hasso Plattner Institute\\
{\tt\small haojin.yang@hpi.de}
\and 
Zhen Shen\footnotemark[1]\\
Alibaba Cloud\\
{\tt\small zackary.sz@alibaba-inc.com}
\and 
Yucheng Zhao\footnotemark[1]\\
ByteDance\\
{\tt\small zhaoyucheng.joe@bytedance.com}
}

\maketitle

\pagestyle{empty}  
\thispagestyle{empty} 

\begin{abstract}
\label{sec:abstract}
Deep convolutional neural networks (CNN) have achieved astonishing results in a large variety of applications.
However, using these models on mobile or embedded devices is difficult due to the limited memory and computation resources.
Recently, the inverted residual block becomes the dominating solution for the architecture design of compact CNNs.
In this work, we comprehensively investigated the existing design concepts, rethink the functional characteristics of two pointwise convolutions in the inverted residuals. 
We propose a novel design, called asymmetrical bottlenecks.
Precisely, we adjust the first pointwise convolution dimension, enrich the information flow by feature reuse, and migrate saved computations to the second pointwise convolution.
Doing so we can further improve the accuracy without increasing the computation overhead.
The asymmetrical bottlenecks can be adopted as a drop-in replacement for the existing CNN blocks.
We can thus create AsymmNet by easily stack those blocks according to proper depth and width conditions.
Extensive experiments demonstrate that our proposed block design is more beneficial than the original inverted residual bottlenecks for mobile networks, especially useful for those ultralight CNNs within the regime of $<$220M MAdds.
Code is available at \href{https://github.com/Spark001/AsymmNet}{https://github.com/Spark001/AsymmNet}

\end{abstract}


\section{Introduction}
\label{sec:introcution}

The recent success of deep \emph{Convolution Neural Networks} (CNN) is like the jewel in the crown of modern AI waves \cite{goodfellow2014generative,he2016deep,ren2015faster}.
However, the current CNN models are heavily relying on high-performance computation hardware, such as GPU and TPU, which are normally deployed in a cloud computing environment.
Thus, the client applications have to transmit user data to the cloud to gain deep CNN models' benefits.
This constraint strongly limits such models' applicability on resource-constrained devices, e.g., mobile phones, IoT devices, and embedded devices.
Moreover, sending user data to a remote server increases the risk of privacy leakage.
Therefore, in recent years, various works aim to solve this problem by reducing memory footprints and accelerating inference.
We roughly categorize those works into following directions:
network pruning \cite{han2015deep,han2015learning}, knowledge distillation \cite{crowley2018moonshine,polino2018model}, low-bit quantization \cite{courbariaux2015binaryconnect,rastegari2016xnor}, and compact network designs \cite{howard2017mobilenets,howard2019searching,sandler2018mobilenetv2,zhang2018shufflenet,ma2018shufflenet,tan2019mnasnet}.
The latter has been recognized as the most popular approach that has a massive impact on industrial applications.
The compact networks achieved promising accuracy with generally fewer parameters and less computation.
Although significant progress has been made, there is still much room for improvement, especially in the ultralight regime with multiply adds (MAdds) $<$220M.

In this paper, our efforts focus on improving ultralight CNNs by the handcrafted design of basic building blocks.
We first made a thorough investigation on existing block designs of mobile CNNs, and we argue that the two pointwise (PW) convolutions contribute differently in the original inverted residual bottleneck block.
\cite{chollet2017xception} first proposes decoupling spatial correlation and channel correlation using the combination of a depthwise (DW) convolution and a PW convolution. 
\cite{sandler2018mobilenetv2} further emphasises that the second PW convolution has essential characteristics in the inverted residual bottlenecks since 
it is responsible for learning new features from different channels, which is especially crucial for expressiveness.
The first PW convolution is responsible for channel expansion, on the other hand.
Therefore, we propose to partially reduce or migrate the first PW computations to the second one.
By following this idea, we introduce a novel \emph{Asymmetrical Bottleneck Block}, as shown in Figure \ref{fig:asymm_detail}.
Furthermore, we can create \emph{AsymmNet} by easily stack a sequence of asymmetrical blocks according to proper depth and width conditions.
We only consider the handcrafted design of the network architecture in this work. However, the proposed CNN block is orthogonal to the recent approaches based on \emph{Neural Architecture Search} (NAS) \cite{tan2019mnasnet,howard2019searching,tan2019efficientnet,radosavovic2020designing}.
Experimental results show that, AsymmNet is especially superior in the ultralight CNN regime, and we have achieved promising results in image classification and other four downstream vision tasks.
Summarized our core contributions in this paper are:
\begin{itemize}
  \item We thoroughly investigated the existing mobile CNN designs and further proposed a novel asymmetrical bottleneck block.
  \item We propose AsymmNet based on asymmetrical bottlenecks, which achieves promising performance under the ultralight CNN regime ($<$220M MAdds) on ImageNet classification and multiple downstream vision tasks.
\end{itemize}

The rest of the paper is organized as follows: 
Section \ref{sec:related_work} briefly review the related work. 
Subsequently, we present the proposed asymmetrical bottleneck design and \emph{AsymmNet} in Section \ref{sec:methodology},
followed by experimental results and discussions (Section \ref{sec:experiment}).
Finally, Section \ref{sec:conclusion} concludes the paper and provides an outlook on future work.

\section{Related Work}
\label{sec:related_work}

This section provides a thorough overview of the recent efforts in the research domain of model compression and compact network design.

In the model compression area, knowledge distillation  \cite{crowley2018moonshine,polino2018model} aims to generate small ``student'' networks trained by using distilled supervision signals derived from a cumbersome ``teacher'' network.
The student network is expected to be more compact and as accurate as of the teacher.
Connection pruning \cite{han2015deep,han2015learning} and channel pruning \cite{li2016pruning,he2017channel} respectively remove low-rank connections between neurons or weakly weighted channels for model shrinking and acceleration.
Low-bit quantization \cite{courbariaux2015binaryconnect,rastegari2016xnor,liu2018bi,liu2020reactnet,bethge2020meliusnet} is another crucial complementary approach to improve network efficiency through reduced precision arithmetic or even bit-wise (binary) operators.
Among them, Bethge et al. \cite{bethge2020meliusnet} introduced a novel block design that suggests applying \emph{DenseNet} \cite{huang2017densely} style concatenation for preserving a rich information flow and a subsequent improvement block to update the newly added features, which can reduce the computation overhead as well.
Our approach is also partially inspired by this concept.

The compact network methods use full precision floating point numbers as weights but reduce the total number of parameters and operations through compact architecture design while minimizing accuracy loss. 
The commonly used techniques include replacing a large portion of 3$\times$3 filters with smaller 1$\times$1 filters \cite{iandola2016squeezenet}; 
Using depthwise separable convolution to reduce operations \cite{chollet2017xception};
Utilizing channel shuffling and group convolutions in addition to depthwise convolution \cite{zhang2018shufflenet}.
Among those approaches, the MobileNet series (V1-V3) \cite{howard2017mobilenets,sandler2018mobilenetv2,howard2019searching} are so far the most successful lightweight CNN models based on depthwise separable convolution and intelligent architecture design.
Specifically, MobileNetV3 combines handcrafted block design and architecture search techniques.
GhostNet \cite{han2020ghostnet} adopts the network architecture of MobileNetV3 but proposed to use a computationally cheaper block design replacing the inverted bottleneck block.
Zhou et al. \cite{zhou2020rethinking} proposed a sandglass block to replace the commonly used inverted bottleneck block, whilst better accuracy can be achieved compared to MobileNetV2 without increasing parameters and computation.

NAS techniques aim to automatically search efficient network architectures \cite{tan2019mnasnet,howard2019searching,tan2019efficientnet,radosavovic2020designing}. 
However, the most efficient basic building block design still requires human expertise \cite{howard2019searching,zhou2020rethinking,han2020ghostnet}.
Furthermore, such methods need to repeat the network design process and retrain the network from scratch for each setting, which will result in excessive energy consumption and $CO_{2}$ emission.
E.g., a \emph{Transformer} language model \cite{vaswani2017attention} with NAS will cause $CO_{2}$ emission as much as 5 cars’ lifetime \cite{strubell2019energy}.


\begin{figure*}[htbp]
\centering
\subfigure[Bottleneck block \cite{he2016deep}]{
\begin{minipage}[t]{0.3\linewidth}
\centering
\includegraphics[width=.9in]{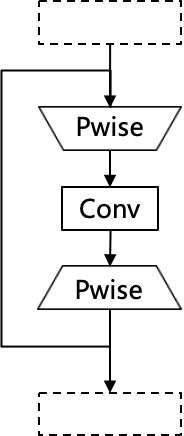}
\label{fig:bottleneck}
\end{minipage}%
}%
\subfigure[Inverted residual block \cite{sandler2018mobilenetv2}]{
\begin{minipage}[t]{0.3\linewidth}
\centering
\includegraphics[width=.9in]{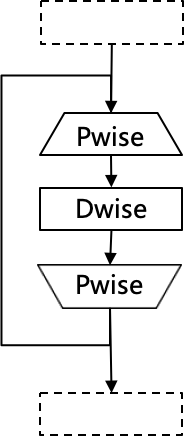}
\label{fig:bottleneck_inverted}
\end{minipage}%
}%
\subfigure[Sandglass block \cite{zhou2020rethinking}]{
\begin{minipage}[t]{0.3\linewidth}
\centering
\includegraphics[width=.97in]{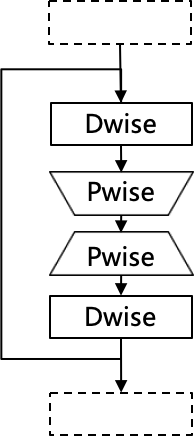}
\label{fig:sandglass}
\end{minipage}
}%
\quad
\subfigure[ShuffleBlock v2 \cite{ma2018shufflenet}]{
\begin{minipage}[t]{0.3\linewidth}
\centering
\includegraphics[width=1in]{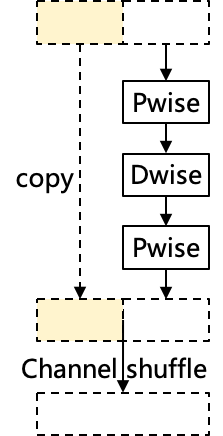}
\label{fig:shufflenet_v2}
\end{minipage}%
}%
\subfigure[Ghost module \cite{han2020ghostnet}]{
\begin{minipage}[t]{0.3\linewidth}
\centering
\includegraphics[width=.85in]{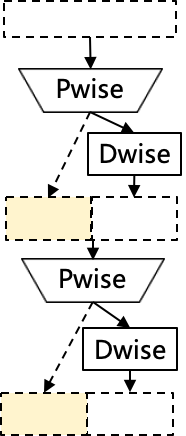}
\label{fig:ghost_module}
\end{minipage}
}%
\subfigure[Asymmetrical bottleneck]{
\begin{minipage}[t]{0.3\linewidth}
\centering
\includegraphics[width=1.05in]{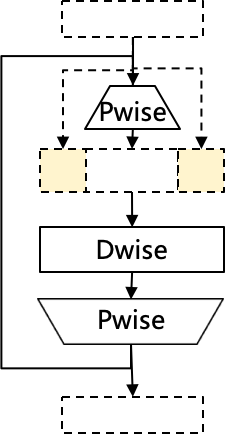}
\label{fig:asym_module}
\end{minipage}
}%
\centering
\caption{Different types of basic convolution blocks. 
``Pwise'' denotes $1\times1$ pointwise convolution, ``Dwise'' denotes $3\times3$ depthwise convolution.
The dotted rectangles and arrows represent feature maps and feature reuse, respectively.}
\label{fig:different_block}
\end{figure*}

\section{Methodology}
\label{sec:methodology}

In this section, we first revisit the existing building block designs for lightweight CNN models.
We then introduce the proposed \emph{asymmetrical bottleneck block} and \emph{AsymmNet}, discuss the design concept and main differences compared to the existing approaches.

\subsection{Preliminaries}

\noindent
\textbf{Depthwise separable convolution}
is proposed by Chollet in the \emph{Xception} network \cite{chollet2017xception}, which is way more efficient than other CNN networks at that time.
Subsequently, this design has been applied in many lightweight CNN architectures such as MobileNet series \cite{howard2017mobilenets,sandler2018mobilenetv2,howard2019searching} and ShuffleNet series \cite{ma2018shufflenet,zhang2018shufflenet}.
It assumes that separately learning spatial and channel correlations should be more efficient and easier for a CNN learner.
Specifically, it replaces a standard convolutional operator by splitting convolution into two separate operators (layers).
The first one is called depthwise convolution, which adopts single channel filters to learn spatial correlations among locations within each channel separately. 
The second operator is a $1\times1$ convolution, also served as pointwise convolution, which is utilized for learning new features through computing linear combinations across all the input channels.
According to \cite{chollet2017xception,howard2017mobilenets}, depthwise separable convolution can reduce computation overhead by a factor of around $kernel\_size^{2}$ compared to a standard convolution operator with the same kernel size.
We also apply depthwise separable convolution in the proposed approach due to its computational efficiency.
\linebreak

\noindent
\textbf{Inverted bottleneck}
is proposed by Sandler et al. for MobileNetV2 \cite{sandler2018mobilenetv2}.
Unlike the original bottleneck design \cite{szegedy2015going} (see Figure \ref{fig:bottleneck}), the inverted bottleneck block adopts a low-dimensional (the number of input channels) input tensor and expands it to a higher dimensional tensor using a pointwise convolution.
The expanded high-dimensional tensor will then be fed into a depthwise separable convolution, by which the corresponding pointwise convolution generates low-dimensional new features by linearly combining channels after a depthwise convolution.
It can be seen that the first pointwise convolution expands the information flow, which increases the capacity, and the subsequent convolution operators are responsible for the expressiveness of the proper layer.
This speculation is derived based on the analysis of the block's capacity and expressiveness in \cite{sandler2018mobilenetv2}.
Figure \ref{fig:bottleneck_inverted} shows the design idea of an inverted bottleneck block.
\linebreak

\begin{table}[ht!]
\begin{center}
\begin{tabular}{c|c|c|c}
\Xhline{4\arrayrulewidth}
Network     & DW (\%) & PW (\%) & Vanilla (\%) \\ \hline
MobileNetV1 &   3.1     & 95      &   1.9     \\
MobileNetV2 &   6.2      & 84.4      &  9.4      \\
MobileNetV3 &   8.9     & 88.5      &  2.6      \\
\Xhline{4\arrayrulewidth}
\end{tabular}
\end{center}
\caption{Computational complexity distribution of MobileNet V1-V3.
We calculate the proportion of MAdds of different types of operators.
DW and PW respectively represent the depthwise and pointwise convolution in the corresponding MobileNet blocks.
Vanilla denotes the computation of the remaining standard convolutions.}
\label{tab:computation_eval_mb}
\end{table}

\noindent
\textbf{Cheap operations for more features}
Table \ref{tab:computation_eval_mb} shows the complexity evaluation results of different types of convolutions of MobileNet series.
We observe that the computation overhead is mainly concentrated on the pointwise convolution part, as e.g., $95\%$ of MobileNetV1, $84.4\%$ of MobileNetV2 and $88.5\%$ of MobileNetV3.
If we want to reduce the computational complexity further, the optimization of this part of the network is the first choice.
Han et al. proposed to use the Ghost module (see Section \ref{sec:revisit}) to replace the pointwise convolution layers and partially remove the depthwise convolution layers (only preserve those for downsampling).
The core idea of the Ghost module is to generate more features using computationally cheaper operators.
Our proposed design is also partially inspired by this concept, and we assume that the two pointwise convolutions contribute differently in the structural point of view.
We thus shift the amount of calculation to the more important one.

\subsection{Revisit existing design}
\label{sec:revisit}
In this section, we review several commonly used design principles: 
original bottleneck block \cite{he2016deep}, inverted residual block \cite{sandler2018mobilenetv2}, shuffleblock v2 \cite{ma2018shufflenet}, Ghost module \cite{han2020ghostnet}, and Sandglass block \cite{zhou2020rethinking}.
Figure \ref{fig:different_block} demonstrates the specific properties of each design.

Bottleneck (Figure \ref{fig:bottleneck}) is the fundamental building block of ResNet \cite{he2016deep}, where a pointwise convolution reduces the feature dimension with the factor $t$, apply a $3\times3$ convolution to the narrowed features, and then utilize another pointwise convolution to restore the feature dimension to be equal to the input size. 
The key difference between Inverted Residual block (also called MMBlock, see Figure \ref{fig:bottleneck_inverted}) and original Bottleneck is that the latter  applies standard convolution on narrowed features, while MMBlock uses the first pointwise convolution to expand the feature dimension, and applies depthwise convolution on expanded features.
It is so because standard $3\times3$ convolutions are highly computational intensive in Bottlenecks.
However, depthwise convolutions in MMBlock can significantly reduce the computational complexity.
Thus, increasing the feature dimension will be beneficial for improving the representative capacity of the block.

Shuffleblock v2 is the following work of Shuffleblock v1 \cite{zhang2018shufflenet}, in which the group convolution is removed for practical efficiency. 
Furthermore, the input feature map in Shuffleblock v2 is split into two equal channels of narrowed feature maps (Figure \ref{fig:shufflenet_v2}). 
One is transformed with a special Bottleneck block without internal dimension changes (the solid arrow path on the right); 
The other (the dashed arrow path on the left) keeps unchanged until concatenated and shuffled with the transformed feature map. 
This design reveals that partially reusing the input features doesn't impair the expressiveness of the convolution blocks, but can effectively reduce computational complexity.

Ghost module (Figure \ref{fig:ghost_module}) is proposed to reduce redundancy of feature maps generated by pointwise convolutions in an MMBlock. 
Specifically, the amount of output channels of pointwise convolutions is reduced to make more room for integrating cheaper intrinsic features. 
To keep the output dimension consistent, a series of linear transformation such as depthwise convolutions is used for generating intrinsic features, which will be concatenated with the output of the pointwise convolution to form the final feature vector.

\cite{zhou2020rethinking} propposed Sandglass block (see Figure \ref{fig:sandglass}), which suggests keeping the standard bottleneck structure.
It prefers to perform identity mapping and spatial transformation at a higher dimension to alleviate information loss and gradient confusion.
Therefore, the sandglass block flips the position of depthwise and pointwise convolutions, which aims to preserve dense information flow and suppress the computation cost.

Based on the previous findings, we argue that improving the capacity by using cheaper intrinsic features or even directly feature reuse is beneficial.
We thus rethink the functional characteristics of two pointwise convolutions in the inverted residuals and propose a novel asymmetrical bottleneck described in the next section.

\subsection{Asymmetrical bottlenecks}
\label{sec:asym_block}

\begin{figure}[htbp]
\centering
\subfigure[Inverted residual block]{
\begin{minipage}[t]{0.5\columnwidth}
\centering
\includegraphics[width=0.95in]{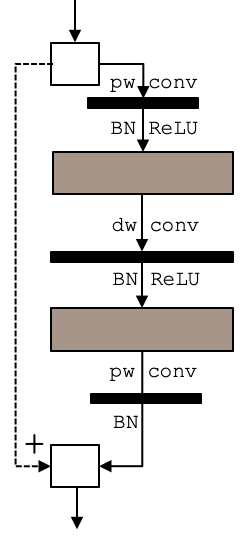}
\label{fig:mmblock}
\end{minipage}%
}%
\subfigure[Pruned block]{
\begin{minipage}[t]{0.5\columnwidth}
\centering
\includegraphics[width=1in]{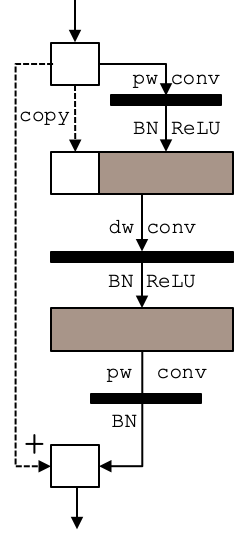}
\label{fig:pruned}
\end{minipage}%
}%
\quad
\subfigure[Asymmetrical block]{
\begin{minipage}[t]{\columnwidth}
\centering
\includegraphics[width=2in]{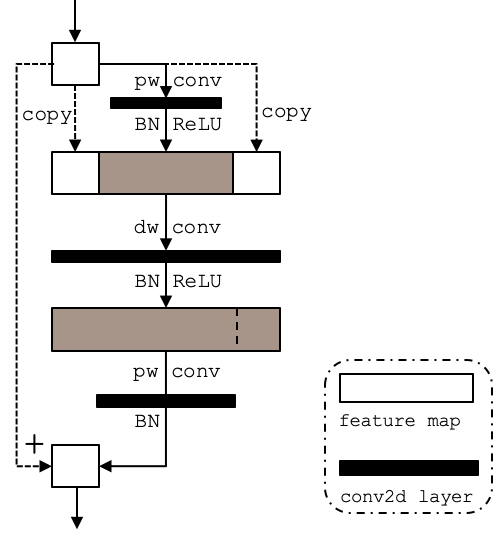}
\label{fig:asymm_detail}
\end{minipage}
}%
\centering
\caption{Detailed illustration of the inverted residual block, pruned block, and asymmetrical bottleneck block.
Brown fillings represent the feature maps generated by convolutions, while white fillings denote feature map reuse.}
\label{fig:block_detail_compare}
\end{figure}


As demonstrated in Table \ref{tab:computation_eval_mb}, pointwise (PW) convolution is the most computationally intensive part in inverted residual bottlenecks (see Figure \ref{fig:mmblock}).
The first PW convolution is adopted to expand the feature tensor's dimension and the second one is significant for learning feature correlations from different channels after the depthwise convolution (DW).
We can figure out that the first PW expands the information flow, which increases the capacity, and the second PW convolution is mainly responsible for the expressiveness.
We argue that cheaper transformations or even feature reuse can enhance the information flow, but learning channel correlations should not be simplified in relative terms.
Therefore, we infer that the second PW has more essential characteristics in the structure.
To verify our speculation, we first designed a pruned version (referred to as pruned block subsequently) based on inverted residual bottlenecks, as shown by Figure \ref{fig:pruned}.
The output of the first PW is expressed by Eq. \ref{eq:mbv3_pw1_output}:
\begin{equation}
\begin{aligned}
    Y_{pw1}&=Concat(X, Y_{t-1}(X))
\end{aligned}
\label{eq:mbv3_pw1_output}
\end{equation}
where $X\in\mathbb{R}^{h \times w \times c}$ denotes the input tensor,
while $h$, $w$, and $c$ denote the height, width, and channel dimension, respectively.
$Y_{t-1}\in\mathbb{R}^{h \times w \times (t-1)\cdot c}$ is the output of the first PW, 
where $t$ is an expansion factor introduced in \cite{sandler2018mobilenetv2}.
In the pruned block, we reduce the output channels of the first PW by $c$.
Thus, the pruned block can be formulated as:
\begin{equation}
\begin{aligned}
    Y_{p}&=X+PW(DW(Concat(X, Y_{t-1}(X))))
\end{aligned}
\label{eq:pruned}
\end{equation}
where we omit ReLU and BatchNorm for simplicity in the formulation.

\begin{table}[]
\footnotesize
\begin{center}
\begin{tabular}{lll}
\Xhline{4\arrayrulewidth}
Input                                         & Operator          & Output                                        \\ \hline
$h \times w \times c$                         & $1\times1$,conv2d,non-linear    & $h \times w \times (t-r)c $                   \\
$h \times w \times (t-r)c $                   & Concat            & $ h \times w \times (t+r)c$                   \\
$h \times w \times (t+r)c $                   & DW(k) s=$s$,non-linear & $\frac{h}{s} \times\frac{w}{s} \times (t+r)c$ \\
$\frac{h}{s} \times\frac{w}{s} \times (t+r)c$ & $1\times1$,conv2d,linear & $\frac{h}{s} \times\frac{w}{s} \times c$      \\
\Xhline{4\arrayrulewidth}
\end{tabular}
\end{center}
\label{tab:basic_description}
\caption{Asymmetrical bottleneck block with stride $s$, asymmetry rate $r$, and expansion factor $t$. }
\end{table}

Pruned block can save computation, but at the same time, it also brings a small amount of accuracy loss.
Therefore, we consider migrating the saved computation of the first PW to the second PW to construct an asymmetrical structure, as shown in Figure \ref{fig:asymm_detail}.
Experimental results show that the performance can be improved with this asymmetrical structure while the computation amount is basically unchanged.
Mathematically, the asymmetrical bottleneck block can be expressed by Eq. \ref{eq:asymm}:
\begin{equation}
\begin{aligned}
    Y'&=X+PW(DW(Concat(2r\cdot X, Y_{t-r}(X)))) 
\end{aligned}
\label{eq:asymm}
\end{equation}
where $Y_{t-r}\in\mathbb{R}^{h \times w \times (t-r)\cdot c}$ is the output of the first PW, $t$ denotes the expansion factor and $r \in [0,t)$ is a new parameter which controls the asymmetry rate.
To achieve a good trade-off between accuracy and efficiency, we set $r$ to 1 in all experiments.
If $r=0$ it degenerates into an inverted residual bottleneck.

\subsubsection{Computational complexity}
Similar to MMBlock, the theoretical computation complexity of AsymmBlock is $C=C_{pw1}+C_{dw}+C_{pw2}$. 
For simplicity, we only calculate the blocks whose $stride=1$. 
So the theoretical complexity ratio of AsymmBlock and MMBlock can be calculated as 
\begin{equation}
\begin{aligned}
R_c &=\frac{hwc(tc-rc)+k^2hw(tc+rc)+hw(tc+rc)c}{hwc(tc)+k^2hw(tc)+hw(tc)c} \\
   &= \frac{2hwtc^2+k^2hwtc+hwrck^2}{2hwtc^2+k^2hwtc} \\
   &= 1+\frac{rk^2}{2tc+k^2t} \approx 1.
\end{aligned}
\end{equation}
where $t$ denotes the expansion factor, $r$ represents the asymmetry rate, and $k$ indicates the kernel size of DW convolution. 
We set $r=1$ in our experiments and $k^2\ll c$. 
$R_c\approx1$ means that the AsymmBlock can transfer the computation cost from the first PW to the second and keep total complexity roughly unchanged.

\subsection{AsymmNet}
\label{sec:asym_net}
\begin{table}[]
\footnotesize
\begin{center}
\begin{tabular}{c|c|c|c|c|c|c}
\Xhline{4\arrayrulewidth}
\multicolumn{1}{l|}{\textit{\textbf{No.}}} & \textbf{Input}                      & \textbf{Operator}      & \textit{\textbf{k}}     & \textit{\textbf{p}} & \textit{\textbf{c}}  & \textit{\textbf{s}} \\ \Xhline{2\arrayrulewidth}
1                                           & $224^{2}\times3$                    & conv2d                    & 3  & -                   & 16                 & 2                   \\ \hline
2                                           & \multirow{2}{*}{$112^{2}\times 16$} & \multirow{2}{*}{asymm-bneck} & 3 & 16                    & 16                 & 1                   \\ 
3                                           &                                     &                              &  3 & 64                  & 24                 & 2                   \\ \hline
4                                           & \multirow{2}{*}{$56^{2}\times 24$}  & \multirow{2}{*}{asymm-bneck} &  3 &    72              & 24                 & 1                   \\ 
5                                           &                                     &                              &  5 &   72               & 40                 & 2                   \\ \hline
6                                           & \multirow{3}{*}{$28^{2}\times 40$}  & \multirow{3}{*}{asymm-bneck} & 5 & 120                  & 40                 & 1                   \\ 
7                                           &                                     &                              & 5 &     120               & 40                 & 1                   \\ 
8                                           &                                     &                              & 3 &      240              & 80                 & 2                   \\ \hline
9                                           & \multirow{4}{*}{$14^{2}\times 80$}  & \multirow{4}{*}{asymm-bneck} & 3 &      200              & 80                 & 1                   \\ 
10                                          &                                     &                              & 3 &      184              & 80             
& 1                   \\ 
11                                          &                                     &                              & 3 &      184             & 80             
& 1                   \\ 
12                                          &                                     &                              & 3 &      480              & 112                & 1                   \\ \hline
13                                          & \multirow{2}{*}{$14^{2}\times 112$} & \multirow{2}{*}{asymm-bneck} & 3 &  672              & 112                & 1                   \\ 
14                                          &                                     &                              & 5 & 672               & 160                & 2                   \\ \hline
15                                          & \multirow{2}{*}{$7^{2}\times 160$}  & \multirow{2}{*}{asymm-bneck} & 5&     960               & 160                & 1                   \\ 
16                                          &                                     &                              & 5&     960               & 160                & 1                   \\ \hline
17                                          &       $7^{2}\times 160$                     & conv2d    & 1      &       -              & 960        & 1                   \\ \hline
18                                          & $7^{2}\times 960$                   & avgpool & 7          & -                    & -                  & 1                   \\ \hline
19                                          & $1 \times 960$                      & conv2d & 1           & -                   & 1280               & 1                   \\ \hline
20                                          & $1 \times 1280$                     & conv2d & 1           & -                   & 1000               & 1                   \\ \Xhline{4\arrayrulewidth}
\end{tabular}
\end{center}
    \caption{Specification for AsymmNet-L using MobileNetV3-large base. Each row shows a conv2d layer or an asymmetrical bottleneck block. $c$ denotes the output channel size, $k$ denotes the kernel size, and $s$ is the stride number of the convolution layer. ``Input'' and ``Operator'' indicate the shape of the input tensor and the operator type. $p$ denotes the expanded channel size of the corresponding asymmetrical bottleneck blocks.}
\label{tab:net_mbv3_large}
\end{table}
\begin{table}[]
\footnotesize
\begin{center}
\begin{tabular}{c|c|c|c|c|c|c}
\Xhline{4\arrayrulewidth}
\multicolumn{1}{l|}{\textit{\textbf{No.}}} & \textbf{Input}                      & \textbf{Operator}      & \textit{\textbf{k}}     & \textit{\textbf{p}} & \textit{\textbf{c}}  & \textit{\textbf{s}} \\ \Xhline{2\arrayrulewidth}
1                                           & $224^{2}\times3$                    & conv2d                    & 3  & -                   & 16                 & 2                   \\ \hline
2                                           & $112^{2}\times 16$                  & asymm-bneck               & 3 & 16                    & 16                 & 2                   \\ \hline
3                                           & $56^{2}\times 16$  & asymm-bneck &  3 &    72              & 24                 & 2                   \\ \hline
4                                           & \multirow{2}{*}{$28^{2}\times 24$}  & \multirow{2}{*}{asymm-bneck} & 3 & 88                  & 24                 & 1                   \\ 
5                                           &                                     &                              & 5 &     96               & 40                 & 2                   \\ \hline
6                                           & \multirow{3}{*}{$14^{2}\times 40$}  & \multirow{3}{*}{asymm-bneck} & 5 &      240              & 40                 & 1                   \\ 
7                                           &                                     &                               & 5 &      240              & 40                 & 1                   \\ 
8                                          &                                     &                              & 5 & 120               & 48                & 1                   \\ \hline
9                                          & \multirow{2}{*}{$14^{2}\times 48$}  &  \multirow{2}{*}{asymm-bneck} & 5 & 144               & 48                & 1                   \\
10                                          &                                     &                              & 5 & 288               & 96                & 2                   \\ \hline
11                                          & \multirow{2}{*}{$7^{2}\times 96$}  & \multirow{2}{*}{asymm-bneck} & \multirow{2}{*}{5}&    \multirow{2}{*}{576}               & \multirow{2}{*}{96}                & \multirow{2}{*}{1}                   \\ 
12                                          &                                     &                              &  &                    &                  &                    \\ \hline
13                                          &       $7^{2}\times 96$                     & conv2d    & 1      &       -              & 576        & 1                   \\ \hline
14                                          & $7^{2}\times 576$                   & avgpool & 7          & -                    & -                  & 1                   \\ \hline
15                                          & $1 \times 576$                      & conv2d & 1           & -                   & 1024               & 1                   \\ \hline
16                                          & $1 \times 1024$                     & conv2d & 1           & -                   & 1000               & 1                   \\ \Xhline{4\arrayrulewidth}
\end{tabular}
\end{center}
    \caption{Specification for AsymmNet-S using MobileNetV3-small base.}
\label{tab:net_mbv3_small}
\end{table}

We further develop several efficient CNN architectures based on the proposed asymmetrical bottleneck design.
To gain the best practical benefits, we follow the basic network architecture of MobileNetV3-large and MobileNetV3-small \cite{howard2019searching}, as shown in Table \ref{tab:net_mbv3_large} and Table \ref{tab:net_mbv3_small}.
The main reason for our choice is that the core hyper-parameters such as kernel size, expand size, and network depth of MobileNetV3 are determined through a NAS algorithm and exhaustive search process.
Our efforts thus mainly focus on the manual improvement of the basic block design.
For a fair comparison, we keep those automatically selected hyper-parameters unchanged.

Therefore, the main building blocks of \emph{AsymmNet} consists of a sequence of stacked asymmetrical bottleneck blocks, which gradually downsample the feature map resolution and increase the channel number to maintain the whole network's information capacity.
We consider the presented architecture in this work as a basic design, while we believe that the automatic architecture and hyper-parameter search methods can further boost the performance.

\section{Experiments}
\label{sec:experiment}

This section presents detailed experimental results.
We first evaluate the model performance on the ImageNet classification task under various complexity (MAdds) settings.
We further validate the generalization ability and effectiveness of the proposed approach on four downstream tasks, including face recognition, action recognition, pose estimation, and object detection.

\subsection{Experiment setup}
We utilize the deep learning framework MXNet \cite{chen2015mxnet} and the off-the-shelf toolbox Gluon-CV \cite{he2019bag} to implement our models. 
We use the standard SGD optimizer for model training with both decay and momentum of 0.9 and the weight decay is 3e-5.
We use the cosine learning rate scheduler with the initial learning rate of 2.6 for eight GPUs.
The corresponding batch size was set to 256.
Without special declaration, we train all the models for 360 epochs, in which five epochs are conducted for a warm-up phase.
Detailed configurations can be found in our open-source codes \footnote{\href{https://github.com/Spark001/AsymmNet}{https://github.com/Spark001/AsymmNet}}.

\subsection{Image classification}

\begin{table*}[ht!]
\begin{center}
\begin{tabular}{c|c|c|c|c|c|c}
\Xhline{4\arrayrulewidth}
Multiplier            &      Model Scale          &  Networks & Top-1 Acc (\%) & MAdds (M) & Params (M) & Latency (ms)      \\ \hline
\multirow{6}{*}{0.35} &  \multirow{3}{*}{Large}   & AsymmNet  & \textbf{65.4}  & 43        & 2.2        &   7.2         \\ 
                      &                           & Pruned    & 63.3           & 36.9      & 2.1        &  \textbf{5.2} \\
                      &                           & MBV3      & 64.2           & 40        & 2.2        &   6.3         \\ \cline{2-7} 
                      &  \multirow{3}{*}{Small}   & AsymmNet  & \textbf{55}    & 15        & 1.7        &   3.3         \\  
                      &                           & Pruned    & 53.2           & 13.6      & 1.7        &   \textbf{2.9}\\
                      &                           & MBV3      & 49.8           & 12        & 1.4        &   3           \\ \hline
\multirow{6}{*}{0.5} &  \multirow{3}{*}{Large}    & AsymmNet  & \textbf{69.2}   & 67.2      & 2.8        &   10.2         \\ 
                      &                           & Pruned    & 68.3           & 59      & 2.6        &  \textbf{7.1} \\
                      &                           & MBV3      & 68.8           & 69        & 2.6        &   8.8         \\ \cline{2-7} 
                      &  \multirow{3}{*}{Small}   & AsymmNet  & \textbf{58.9}  & 20.6        & 1.9        &   4.2         \\  
                      &                           & Pruned    & 57.3           & 18.6      & 1.9        &   \textbf{3.6}\\
                      &                           & MBV3      & 58             & 21        & 1.6        &   3.7           \\ \hline
\multirow{6}{*}{0.75} &  \multirow{3}{*}{Large}   & AsymmNet  & \textbf{73.5}  & 142.1      & 4.2        &   19.4         \\ 
                      &                           & Pruned    & 72.6           & 125.3      & 3.8        &  \textbf{13.6} \\
                      &                           & MBV3      & 73.3           & 155        & 4          &   16.2         \\ \cline{2-7} 
                      &  \multirow{3}{*}{Small}   & AsymmNet  & \textbf{65.6}  & 40.8       & 2.5        &   6.9         \\  
                      &                           & Pruned    & 64             & 36.9       & 2.3        &   \textbf{6.1}\\
                      &                           & MBV3      & 65.4           & 44         & 2          &   6.3           \\ \hline
\multirow{6}{*}{1.0} &  \multirow{3}{*}{Large}   & AsymmNet  & \textbf{75.4}  & 216.9        & 5.99        &   27.1         \\ 
                      &                           & Pruned    & 74.9           & 193.6      & 5.3        &  \textbf{19.5} \\
                      &                           & MBV3      & 75.2           & 216.5        & 5.4        &   23.3         \\ \cline{2-7} 
                      &  \multirow{3}{*}{Small}   & AsymmNet  & \textbf{68.4}    & 57.7        & 3.1        &   8.9         \\  
                      &                           & Pruned    & 67           & 52.5      & 2.6        &   \textbf{7.9}\\
                      &                           & MBV3      & 67.5           & 57        & 2.5        &   8.17           \\ \hline
\multirow{6}{*}{1.25} &  \multirow{3}{*}{Large}   & AsymmNet  & 76.4           & 349.8     & 8.3        &   38.8         \\ 
                      &                           & Pruned    & 76.1           & 311       & 7.2        &  \textbf{29.2} \\
                      &                           & MBV3      & \textbf{76.6}  & 356       & 7.5        &   34.9         \\ \cline{2-7} 
                      &  \multirow{3}{*}{Small}   & AsymmNet  & \textbf{70.6}    & 91.7        & 3.9        &   12.7         \\  
                      &                           & Pruned    & 69.8           & 83.1      & 3.5        &   \textbf{11.2}\\
                      &                           & MBV3      & 70.4           & 91        & 3.6        &   11.7           \\ \hline
\Xhline{4\arrayrulewidth}
\end{tabular}
\end{center}
\caption{Performance comparison between AsymmNet, Pruned model, and MobileNetV3 across a large variety of scale levels.
We consider both V3-Large and V3-small architecture as references and comprehensively evaluated the accuracy, computation complexity (MAdds), and inference efficiency.
Top-1 accuracy is on the ImageNet dataset, and all latency are obtained by averaging the inference time of 1000 executions on a Qualcomm snapdragon-855 CPU with 8G RAM in the single thread modus.
To conduct a fair comparison, we replace the corresponding convolution blocks and keep all the hyper-parameters unchanged.}
\label{tab:imagenet_compare_mbv3l}
\end{table*}

\begin{table}
\footnotesize
\begin{center}
\begin{tabular}{c|c|c|c|c}
\Xhline{4\arrayrulewidth}
Scale                        & $r$          & Top-1 Acc (\%) & MAdds (M) & Params (M)  \\ \hline
\multirow{3}{*}{Large}       & 0            & 75.2           & 216.6    & 5.4                  \\ 
                             & 1            & \textbf{75.4}           & 216.9    & 5.9                  \\  
                             & 2            & 74.8           & 217.3    & 6.6                 \\  \hline
\multirow{3}{*}{Small}       & 0            & 67.4           & 56.9     & 2.9                   \\  
                             & 1            & \textbf{68.4}           & 57.7     & 3.1                   \\  
                             & 2            & 68.0           & 58.5     & 3.3                   \\ \Xhline{4\arrayrulewidth}
\end{tabular}
\end{center}
\caption{Ablation study on asymmetry rate $r$ using ImageNet dataset.
We apply both large and small AsymmNet for this evaluation.
$r=1$ shows the best performance.}
\label{tab:imagenet_ablation}
\end{table}

\subsubsection{Compare to MobileNetV3}
This section extensively studies the advantages of the proposed AsymmNet and pruned model over MobileNetV3 (MBV3) on the ImageNet ILSVRC 2012 dataset \cite{deng2009imagenet}.
As shown in Table \ref{tab:imagenet_compare_mbv3l}, we compare their performance under multiple complexity settings by tuning the weight multiplier.
We consider both V3-large and V3-small architecture as references and comprehensively evaluated the classification accuracy, computation complexity (MAdds), and inference latency.
Doing so can help reveal the performance advantage of the full spectrum of model architecture configurations.
We applied the DNN inference toolkit MNN \cite{alibaba2020mnn} for the latency evaluation since MNN is specifically optimized for mobile devices. 
All the inference tests have been done on an Android phone equipped with a Qualcomm snapdragon-855 CPU with 8G RAM in the single thread modus.
We average the latency results of 1000 runs for each model.
To conduct a fair comparison, we replace the corresponding convolution blocks and keep all the hyper-parameters unchanged.

We can figure out that, AsymmNet outperforms MobileNetV3 on classification accuracy in almost all the complexity settings, 
while the pruned model demonstrates better efficiency with slight accuracy drops.
However, the accuracy loss becomes negligible when the MAdds getting smaller or even reversed, e.g., pruned model outperforms MobileNetV3 by 3.4\% at the level of 0.35-Small.
Specifically, when the model gets smaller and less complex, the accuracy advantage of AsymmNet becomes more apparent.
This phenomenon effectively reveals the proposed asymmetrical bottleneck block's superiority in the spectrum of extremely lightweight models (a regime $<$220M MAdds).

\subsubsection{Ablation study on asymmetry rate}

We evaluate the asymmetry rate $r$ on the ImageNet dataset to obtain the best choice in terms of accuracy and complexity.
Table \ref{tab:imagenet_ablation} shows the result, where $r=1$ demonstrates the best trade-off.

\subsection{Face recognition}
\begin{table*}[htbp]
\begin{center}
\small
\begin{tabular}{c|c|c|c|c|c|c|c}
\Xhline{4\arrayrulewidth}
Backbone      & LFW (\%)       & CALFW (\%)     & CPLFW (\%)     & AgeDB-30 (\%)  & VGGFace2 (\%)    & MAdds(M) &  Params(M) \\ \hline
AsymmNet-s    & \textbf{97.9} & \textbf{90.1} & \textbf{80.2} & \textbf{86.6} & \textbf{83.6}   & 10.3    &    0.3 \\
MBV3-s        &         97.2  & \textbf{90.1}  &         78.7  &         85.7  &         83.5    & 10.2    &   0.3 \\ \hline
AsymmNet-L    & \textbf{99.1} &         93.8  & \textbf{86.6} & \textbf{93.2} & \textbf{89.6}   & 41.6    &  1.1 \\
MBV3-L        & \textbf{99.1} & \textbf{94.0}&         84.7  &         93.0  &         88.4     & 41.2    &   1.0 \\ \hline
MobileFaceNet\cite{chen2018mobilefacenets} &         99.6  &         -      &         -      &         -      &         -        & 221      &   1.0 \\
\Xhline{4\arrayrulewidth}
\end{tabular}
\end{center}
\caption{Performance comparison among different face recognition datasets.}
\label{tab:face_recognition_res}
\end{table*}

Face recognition is a crucial identity authentication technology used in many mobile or embedded applications such as mobile payment and device unlock.
In this subsection, we employ AsymmNet-L and the proposed AsymmNet-s as network backbone for face recognition.
Following MobileFaceNet \cite{chen2018mobilefacenets}, we use a global depthwise convolution layer rather than a global average pooling layer to output discriminative feature vectors.
Both models are trained on MS-Celeb-1M \cite{guo2016ms} dataset from scratch by ArcFace \cite{deng2019arcface} loss, for a fair comparison between them.
The input image size is $112 \times 112$.
We report result on different dataset including LFW \cite{huang2008labeled}, CALFW \cite{zheng2017cross}, CPLFW \cite{zheng2018cross}, AgeDB-30 \cite{moschoglou2017agedb}, and VGGFace2 \cite{cao2018vggface2} as in Table \ref{tab:face_recognition_res}.
As shown in the table, our face recognition models with AsymmNet-s/L as backbone outperform MBV3-s/L consistently, especially in the CPLFW dataset, our models outperform by a margin of 1.5\% and 1.9\%, respectively.
The computational complexity of our face recognition model uses AsymmNet-s as the backbone is about 21 times less than MobileFaceNet \cite{chen2018mobilefacenets}.

\subsection{Action recognition}
\begin{table}[]
\small
\begin{center}
\begin{tabular}{c|c|c|c}
\Xhline{4\arrayrulewidth}
Backbone                            &   Top1-Acc (\%)  & MAdds(M) & Params(M) \\ \hline
AsymmNet-s                          &   \textbf{47.2}  & 56.5     & 1.9       \\
MBV3-s                              &           46.7   & 55.7     & 1.7       \\ \hline
AsymmNet-L                          &   \textbf{52.8}  & 215.7    & 4.8       \\
MBV3-L                              &           51.3   & 215.4    & 4.2       \\\hline
$ResNet50_{v1b}$\cite{guo2020gluoncv} &           55.2   & 4087.2   & 16.1      \\
\Xhline{4\arrayrulewidth}
\end{tabular}
\end{center}
\caption{Action recognition result on HMDB51 dataset.}
\label{tab:action_recognition_res}
\end{table}

Action recognition has drawn a significant amount of attention from the academic community, owing to its applications in many areas like security and behavior analysis.
We evaluate and compare AsymmNet and MobileNetV3 as feature extractors for action recognition following \cite{wang2016temporal} on the HMDB51 dataset \cite{Kuehne2011HMDB}.
The input image size is $224 \times 224$.
As summarized in Table \ref{tab:action_recognition_res}, AsymmNet achieves superior results compared to MobileNetV3 at both large and small scales.
Furthermore, it reaches an accuracy close to ResNet50$_{v1b}$ \cite{guo2020gluoncv}, while its MAdds is about 19 times smaller.

\subsection{Pose estimation}
\begin{table}[]
\footnotesize
\begin{center}
\begin{tabular}{c|c|c|c|c|c}
\Xhline{4\arrayrulewidth}
Backbone                          & AP             & $AP_{50}$       & $AP_{75}$         & MAdds(M)  & Params(M)   \\ \hline
AsymmNet-s                        & \textbf{54.4}  & \textbf{84.4}   &  \textbf{59.7}  & 310.7     & 1.7   \\
MBV3-s\cite{guo2020gluoncv}       &         54.3   &         83.7    &          59.4   & 309.7     & 1.6   \\ \hline
AsymmNet-L                        &         63.5   & \textbf{88.9}   &  \textbf{70.9}  & 523.9     & 4.3   \\
MBV3-L\cite{guo2020gluoncv}       & \textbf{63.7}  & \textbf{88.9}   &          70.8   & 523.6     & 3.7   \\
\Xhline{4\arrayrulewidth}
\end{tabular}
\end{center}
\caption{Pose estimation results on COCO human pose dataset using SimplePose method. All the AP results are in percentage.}
\label{tab:pose_estimation_res}
\end{table}

There has been significant progress in pose estimation and increasing interest in pose tracking in recent years.
Thus, we evaluated AsymmNet as the backbone on this task by using the challenging COCO human pose benchmark \cite{lin2014microsoft}.
Our approach is based on the \emph{SimplePose} model \cite{xiao2018simple}, which estimates heat maps from deep and low-resolution feature maps.
We replace the backbone with mobile CNN models and evaluate the accuracy for a fair comparison.
The input image size is $256 \times 192$. 
The test set results are given in Table \ref{tab:pose_estimation_res}. 
Our pose estimation results with AsymmNet-s backbone surpass MBV3-s in all three metrics.

\subsection{Object detection}

\begin{table}[]
\small
\begin{center}
\begin{tabular}{c|c|c|c}
\Xhline{4\arrayrulewidth}
Backbone                         & mAP (\%)          & MAdds(G)    & Params(M) \\ \hline
AsymmNet-s                       &  \textbf{69.80}   & 7.99        & 9.33      \\
MBV3-s                           &           68.98   & 7.99        & 9.16      \\ \hline
AsymmNet-L                       &           76.18   & 8.58        & 11.97      \\
MBV3-L                           &   \textbf{76.64}  & 8.58        & 11.39       \\ \hline
MobileNetV1\cite{guo2020gluoncv} &           75.8    & 9.92        & 11.83       \\
\Xhline{4\arrayrulewidth}
\end{tabular}
\end{center}
\caption{Object detection results on the PASCAL VOC dataset using YOLO-v3 detector.}
\label{tab:object_detetion_res}
\end{table}

We apply AsymmNet as a drop-in replacement for the original backbone in YOLO-v3 detector \cite{redmon2018yolov3}.
We compare our results to MobileNetV3 on the PASCAL VOC dataset \cite{everingham2010pascal} on object detection.
We change the base model of the adopted YOLO-v3 architecture and train our models on the combination of VOC2007 trainval and VOC2012 trainval, test on VOC2007 test set.
The input image size is $416 \times 416$.
Table \ref{tab:object_detetion_res} illustrates the results compared to other models based on MobileNet backbones.
AsymmNet-s outperforms MBV3-s on this task.

\section{Conclusion and Discussion}
\label{sec:conclusion}
In this paper, we introduced a novel design for ultralight CNN models.
We investigated important design choices, redesigned two pointwise convolutions of the inverted residual block, and developed a novel asymmetrical bottleneck.
We can see that AsymmNet has a consistent accuracy advantage over MobileNetV3 in the ultralight model regime through our experiments on a series of downstream tasks.
It thus can be used as a practical complementary approach to existing state-of-the-art CNN models in the regime of $<$220M MAdds.

However, we observed that AsymmBlock also has its limitations. 
For instance, with the increase of MAdds, it can not continue to show the advantage of accuracy, as the comparison result with MobileNetV3-1.25. 
Also, the AsymmNet-L model does not demonstrate benefits in the object detection task.
One possible explanation is that the current AsymmNet architecture is based on MBV3, 
which is searched using MMBlock that not necessarily the most suitable architecture for AsymmNet.
Thus, we will continue to optimize it in future work.
As the next step, we will combine automatic search techniques with asymmetrical bottlenecks.


\section{Acknowledgment}
We would like to thank Chao Qian for his valuable technical support on inference speed evaluation.

{\small
\bibliographystyle{ieee_fullname}
\bibliography{sonic_paper}
}

\end{document}